\documentclass[letterpaper, 10 pt, conference]{ieeeconf}

\IEEEoverridecommandlockouts                  
\usepackage{graphicx}
\usepackage{bm}
\usepackage[utf8]{inputenc}
\usepackage{makecell}
\usepackage{booktabs}
\usepackage{amsmath}
\usepackage{amsfonts}
\usepackage{acronym}
\usepackage[misc]{ifsym}
\usepackage{float}
\usepackage{amssymb}
\usepackage{graphicx}
\usepackage[colorlinks=true, linkcolor=blue, citecolor=blue]{hyperref}
\usepackage{cleveref}
\usepackage{afterpage}   
\usepackage{stfloats}     
\usepackage[absolute,overlay]{textpos}
\usepackage{tabularx}
\usepackage{ragged2e}
\usepackage{siunitx}

\usepackage{xspace}

\acrodef{gad}[GAD]{Generative Adversarial Distillation}
\acrodef{gail}[GAIL]{Generative Adversarial Imitation Learning}
\newcommand{\method}[0]{\texttt{StyleLoco}\xspace}

\title{\LARGE \bf
\method: Generative Adversarial Distillation \\ for Natural Humanoid Robot Locomotion
}

\author{Le Ma$^{1*}$, Ziyu Meng$^{1,2*}$, Tengyu Liu$^{1}$, Yuhan Li$^{1,3}$, Ran Song$^{2}$, Wei Zhang$^{2}$, Siyuan Huang$^{1,~\textrm{\Letter}}$% <-this % stops a space
\vspace{3pt}\\
\small $^1$ National Key Laboratory of General Artificial Intelligence, BIGAI\quad{}
$^2$ School of Control Science and Engineering, Shandong University\\
\small $^3$ Huazhong University of Science and Technology\quad{}
\small $^\star{}$Equal contributors \quad $\textrm{\Letter}$\,\,\texttt{huangsiyuan@bigai.ai}
\vspace{6pt}\\
    \href{https://styleloco.github.io/}{https://styleloco.github.io/}
\vspace{-12pt}
}

\begin{document}

\maketitle

\thispagestyle{empty}
\pagestyle{empty}

%%%%%%%%%%%%%%%%%%%%%%%%%%%%%%%%%%%%%%%%%%%%%%%%%%%%%%%%%%%%%%%%%%%%%%%%%%%%%%%%
\begin{abstract}

Humanoid robots are anticipated to acquire a wide range of locomotion capabilities while ensuring natural movement across varying speeds and terrains. Existing methods encounter a fundamental dilemma in learning humanoid locomotion: reinforcement learning with handcrafted rewards can achieve agile locomotion but produces unnatural gaits, while \ac{gail} with motion capture data yields natural movements but suffers from unstable training processes and restricted agility. Integrating these approaches proves challenging due to the inherent heterogeneity between expert policies and human motion datasets. To address this, we introduce \method, a novel two-stage framework that bridges this gap through a \ac{gad} process. Our framework begins by training a teacher policy using reinforcement learning to achieve agile and dynamic locomotion. It then employs a multi-discriminator architecture, where distinct discriminators concurrently extract skills from both the teacher policy and motion capture data. This approach effectively combines the agility of reinforcement learning with the natural fluidity of human-like movements while mitigating the instability issues commonly associated with adversarial training. Through extensive simulation and real-world experiments, we demonstrate that \method enables humanoid robots to perform diverse locomotion tasks with the precision of expertly trained policies and the natural aesthetics of human motion, successfully transferring styles across different movement types while maintaining stable locomotion across a broad spectrum of command inputs.

\end{abstract}

%%%%%%%%%%%%%%%%%%%%%%%%%%%%%%%%%%%%%%%%%%%%%%%%%%%%%%%%%%%%%%%%%%%%%%%%%%%%%%%%
\section{INTRODUCTION}

Natural and agile locomotion in humanoid robots represents a fundamental challenge in robotics, with far-reaching implications for human-robot interaction, disaster response, and industrial applications. While humanoid robots offer unprecedented potential for operating in human-centric environments, achieving human-like movement patterns remains difficult due to their high degrees of freedom and inherently unstable dynamics\cite{10035484}. This challenge is further complicated by the fundamental trade-off between achieving precise control and maintaining natural motion qualities.

\begin{figure}[t!] % 使用[t!]强制顶部对齐
  \includegraphics[width=\columnwidth]{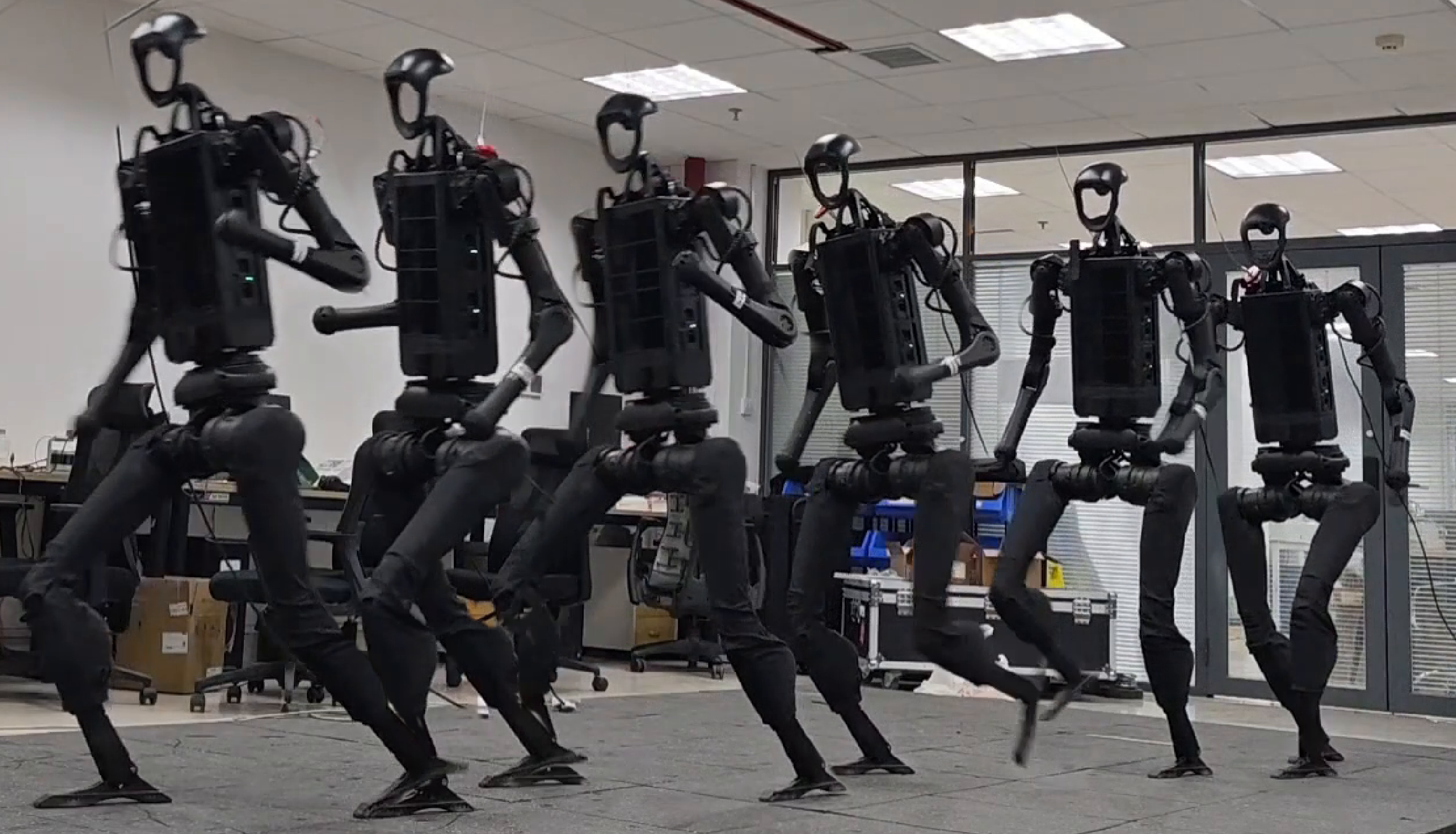}
  \caption{Gait pattern transitions during forward velocity ($v_{x}$) acceleration from 0.7 m/s to 1.8 m/s}
  \label{figs:gait_transition}
\end{figure}

Reinforcement learning (RL) has emerged as a powerful approach for developing locomotion controllers, enabling robots to master complex movements through carefully designed reward functions. These methods often employ a two-stage learning process: first training a teacher policy that relies on privileged information (such as global positions and ground truth environmental parameters) unavailable in real-world settings, then distilling this knowledge into a student policy that operates solely on realistic sensor observations. While this approach has demonstrated impressive results in terms of agility and precision, it faces two key limitations. First, the reliance on handcrafted rewards requires extensive tuning to accommodate different gaits, stride lengths, and motion parameters across varying speeds. Second, these methods often result in rigid, mechanical movements that lack the fluidity and naturalness characteristic of human motion, limiting their effectiveness in human-centric environments.

Recent advances in generative adversarial imitation learning, particularly approaches like Adversarial Motion Prior (AMP)~\cite{peng2021amp}, have opened new possibilities for achieving more natural robot movements by leveraging large-scale motion capture datasets such as LaFAN1~\cite{harvey2020robust} and AMASS~\cite{AMASS:ICCV:2019}. These methods employ adversarial training to ensure that robot movements closely match the statistical patterns present in human demonstrations~\cite{cheng2024expressive}. However, their performance is fundamentally limited by the content and quality of the reference motion data. For instance, learning running behaviors becomes impossible with a dataset containing only walking motions, and acquiring diverse specialized skills often requires expensive motion capture sessions. Furthermore, these methods struggle when motion datasets lack diversity or when retargeting processes introduce artifacts, resulting in brittle behaviors that fail to generalize beyond demonstrated movements.

The limitations of both approaches highlight a critical gap in humanoid locomotion: the need to combine the precision and adaptability of RL-based controllers with the natural movement qualities captured in human demonstrations. While RL methods can learn complex skills beyond available motion capture data, they struggle with natural movement generation. Conversely, demonstration-based methods excel at producing natural movements but are constrained by the available motion capture data. This complementary nature suggests the potential for combining both approaches, yet traditional methods struggle to bridge this gap due to the fundamental heterogeneity between expert policies trained with handcrafted rewards and the statistical patterns present in human motion datasets.

We address these challenges with \method, introducing a novel Generative Adversarial Distillation (GAD) framework that effectively combines knowledge from heterogeneous sources. Our approach employs a multi-discriminator architecture where separate discriminators simultaneously distill skills from both an RL-trained expert policy and motion capture demonstrations. This design allows the model to preserve the agility and precision of RL while incorporating the natural style of human movements, enabling natural skill execution even for behaviors not present in the motion capture data. Through extensive evaluations in both simulated and real-world environments, we demonstrate that \method enables humanoid robots to achieve superior locomotion performance compared to traditional approaches while maintaining natural, human-like movement qualities.

The key contribution of our work is three-fold.
\begin{itemize}
    \item A novel GAD framework that enables stable policy distillation from heterogeneous sources, effectively bridging the gap between RL and demonstration-based approaches.
    \item A multi-discriminator architecture that successfully combines task-oriented control objectives with natural motion patterns, achieving both high performance and human-like movement qualities.
    \item Comprehensive validation through real-world deployment on the Unitree H1 humanoid robot, demonstrating robust and natural motion across diverse locomotion tasks and speeds.
\end{itemize}

\section{RELATED WORKS}

\subsection{Humanoid Robot Locomotion}

Locomotion is a critical aspect in the motion control in humanoid robots.
Traditional methods typically achieve stable movement by formulating the robot's dynamics model as constrained trajectory optimization problems \cite{7442110}. Model Predictive Control (MPC) \cite{romualdi2022online,englsberger2020mptc,elobaid2023online} is then employed in real-time to adjust and execute this trajectory, enabling adaption to dynamic environmental changes. However, these model-based methods usually rely heavily on precise modeling of robot dynamic properties \cite{8461207,9158331,ramos2019dynamic,8063558,6907357} and environmental conditions \cite{MontecilloPuente2010OnRW,5686312,7803405,8246911,ramos2019dynamic,8624943,6907261,7258356,10380694}, which leads to vulnerabilities in real-world performance, 
especially when there is a substantial discrepancy between the applied environments and the predefined conditions \cite{li2024reinforcement}. 
Thus, the optimization problem for humanoid robots is slow to resolve due to the complexity of high-dimensional state and action spaces, rendering it challenging to satisfy the demands for real-time performance and stability.

Recently, reinforcement learning (RL) has emerged as a promising paradigm for humanoid locomotion tasks. These methods design tailored reward functions to guide ``try and error" feedback-based learning process. For instance, reward functions are often crafted to encourage stable walking, minimize energy consumption, or optimize trajectory tracking \cite{fuminimizing}. 
However, designing effective reward functions is non-trivial and often requires extensive domain expertise especially for particular locomotion gaits. 
Natural locomotion motions require different gaits for varying movement speeds, making the design of the reward function even more challenging.
Moreover, the numerous rewards terms must strike a delicate balance between competing objectives. To alleviate these drawbacks, we incorporate diverse reference locomotion motions as style guidance to simplify the reward components and encourage the policy learn versatile gaits.

\subsection{Imitation Learning for Humanoid locomotion}

The fundamental challenges in learning high-dimensional, underactuated robotic systems include precise task specification and effective exploration.
Imitation learning (IL) is a method that learns from expert demonstrations, effectively addressing challenges related to quantifying rewards. Unlike pure reinforcement learning, IL can directly leverage offline expert data to guide policy learning, significantly reducing the exploration space and obtaining dense rewards. This approach is particularly effective in real-world robotics and complex task scenarios. Typically, it involves directly following reference trajectories through motion tracking.
Generative Adversarial Imitation Learning (GAIL) \cite{ho2016generative} has been applied to locomotion tasks. The traditional imitation learning method, as mentioned above, is limited in flexibility—it can only replicate reference trajectories and cannot adapt to downstream tasks. To address this limitation, AMP \cite{peng2021amp} introduces the concept of learning the style from reference motion as a constraint, guiding the policy learning process.

However, this paradigm heavily relies on expert demonstrations, and its performance can significantly degrade when the quality of demonstrations is poor or when the task changes. Since IL strategies are directly derived from the demonstrations, they are prone to overfitting to the demonstration data. As a result, when faced with novel situations, IL may lack sufficient generalization ability.
Furthermore, due to the morphological differences between humanoid robots and humans, obtaining high-quality reference data proves challenging, resulting in datasets that can only encompass a limited range of instructions. This scarcity of data can compromise the stability of Generative Adversarial Imitation Learning (GAIL), leading to mode collapse. To mitigate these challenges, we supplement the expert policy as a reference motion, providing additional motion references to achieve a stable omnidirectional movement strategy.

\subsection{Deployable Policy Distillation}

In robotic locomotion control, distillation is a method that transfers knowledge from teacher policies with privileged information (e.g., full-state dynamics, simulated ground-truth forces, or ideal state estimators) to student policies for real-world deployment. This knowledge transfer enables the student to leverage the teacher’s expertise while operating under real-world constraints, such as partial observation or limited sensory inputs. There are two main approaches to distillation: 

% Distillation based on Behavior Cloning (BC). 
BC methods\cite{huang2024diffuseloco,10610286} learn by mimicking the teacher’s actions using supervised learning on state-action pairs. BC achieves effective performance when the student operates within the teacher’s training distribution, as it directly replicates the teacher’s behavior under familiar conditions. However, its performance degrades sharply with ``compounding error" \cite{pmlr-v9-ross10a} in out-of-distribution (OOD) scenarios (e.g., environmental perturbations, actuator noise, or unseen terrains), as BC inherently lacks the capacity to self-correct deviations from the teacher’s demonstration space. This limitation arises because BC relies solely on static datasets of teacher demonstrations, without mechanisms to adapt to novel or unexpected situations.

Another popular approach is online distillation via Dataset Aggregation (DAgger) \cite{ross2011reduction}, which addresses BC’s limitations by iteratively aggregating student-generated trajectories with teacher-corrected actions. Recently, DAgger and its derivative strategies have stood out as a promising distillation approach for humanoid robot \cite{ji2024exbody2,he2024learning,he2024hover,he2024omnih2o} to acquire deployable policies. During training, the student policy interacts with the environment, while the teacher provides corrective feedback on the student’s actions, enabling the student to refine its policy over multiple iterations. This interactive process mitigates distributional shift and improves robustness to OOD scenarios. However, DAgger still faces a fundamental challenge: the student lacks access to the teacher’s privileged information (e.g., simulated contact forces, ideal state estimators, or full-state dynamics). As a result, under partial observation or incomplete environmental feedback, the student struggles to fully replicate the teacher’s actions.
\cite{fuminimizing}

\section{METHOD}

\method is a novel approach for learning deployable natural locomotion skills that effectively combines the precision of RL-based controllers with the naturalness of human demonstrations. At its core, \method employs our proposed Generative Adversarial Distillation (GAD) framework, which uses a unique double-discriminator architecture to distill knowledge from both an RL-trained teacher policy and human motion demonstrations into a deployable student policy. Through adversarial learning, our approach generates naturalistic motions beyond the constraints of available motion capture data while avoiding the artificial behaviors typically resulting from hand-crafted rewards.

\method consists of three key components: (1) a teacher policy trained with privileged information to achieve robust omnidirectional locomotion, (2) a motion dataset containing natural human movements, and (3) our novel GAD framework that combines these sources to train a deployable student policy. The framework's innovation lies in its ability to generate natural behaviors beyond what either source can achieve alone - overcoming both the limited coverage of motion datasets and the unnatural movements that emerge from pure RL training.

To achieve this, \method employs two discriminators that work in concert to adversarially shape the student policy's behavior. One discriminator ensures the policy can replicate the robust performance of the teacher, while the other maintains consistency with natural human motion patterns. This dual-discriminator approach simultaneously serves two purposes: expanding the range of natural behaviors beyond the demonstration data, and distilling the teacher's capabilities into a deployable policy. The resulting system produces controllers that are both highly capable and naturally moving, without being constrained to demonstrated behaviors or exhibiting artifacts from hand-crafted rewards.

\subsection{Preliminaries}
\subsubsection{Reinforcement Learning}

We formulate humanoid locomotion control as a Partially Observable Markov Decision Process (POMDP) defined by tuple $\langle\mathcal{S},\mathcal{A},T,\mathcal{O},R,\gamma\rangle$, where $\mathcal{S}$ represents the full state space, $\mathcal{O}$ denotes partial observations available to the robot, $\mathcal{A}$ is the action space, $T(s^{\prime}|s,a)$ describes state transitions, $R(s,a)$ defines the reward function, and $\gamma \in (0, 1]$ is the discount factor. The goal is to learn a policy $\pi(a|o)$ that maximizes expected discounted returns while operating only on partial observations $o \in \mathcal{O}$.

The locomotion task requires tracking commanded velocities $v^* = (v^*_x, v^*_y, \omega^*_z)$, where $(v^*_x, v^*_y)$ specify desired linear velocities in local coordinate frame and $\omega^*_z$ defines the desired yaw rate. Following \cite{gu2024advancing}, we use the reward function:
$$r_\textrm{task}(e,\lambda):=\exp(-\lambda \cdot\|e\|^{2})$$
where $e$ represents tracking errors and $\lambda$ controls their relative importance.

\subsubsection{Generative Adversarial Imitation Learning}

Generative Adversarial Imitation Learning (GAIL) learns to mimic expert behavior through adversarial training. Given a dataset of expert demonstrations $\mathcal{M} = {(s_i, a_i)}$ consisting of state-action pairs, GAIL trains a policy $\pi(a|s)$ that generates actions $a'$ for given states $s'$. A discriminator network $\mathcal{D}$ is employed to distinguish between state-action pairs $(s, a)$ from the expert demonstrations and those produced by the policy $\pi$. The reward function used to train the policy is then given by:
$$r_\textrm{GAIL}(s,a)=-\mathrm{log}\left(1-\mathcal{D}(s,a)\right)$$

Adversarial Motion Prior (AMP)~\cite{peng2021amp} extends this framework to handle settings where only state information is available in the demonstrations. Instead of operating on state-action pairs, AMP's discriminator evaluates state transitions $(s, s')$, enabling imitation learning from state-only demonstrations. Additionally, AMP employs a least-squares discriminator~\cite{mao2017least}, replacing the traditional binary cross-entropy loss, which has been empirically shown to provide more stable adversarial training dynamics.

\subsection{Generative Adversarial Distillation}

\begin{figure}[t]
    \centering
    \includegraphics[width=0.48\textwidth]{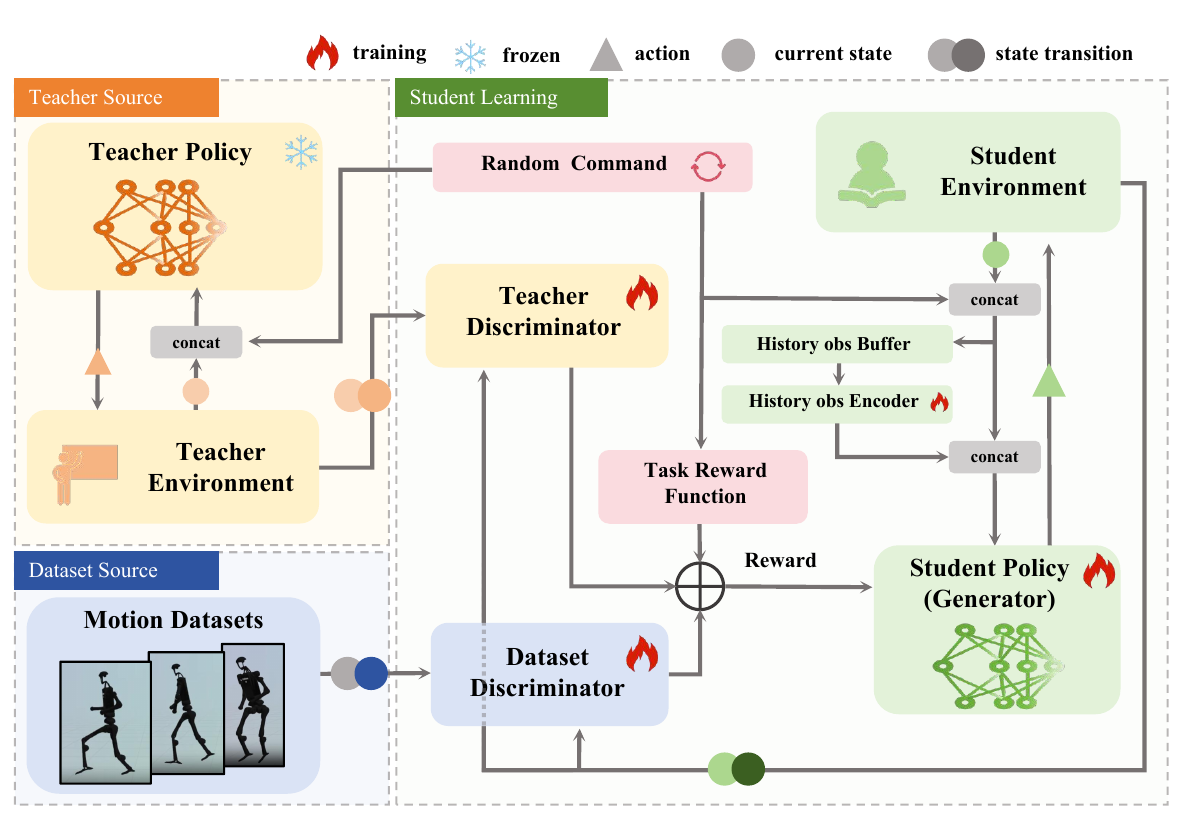}
    \caption{Overview of the proposed Generative Adversarial Distillation (GAD) framework. Two discriminators separately evaluate the similarity of generated motions against a teacher policy and reference motion dataset, enabling the synthesis of natural and adaptive behaviors.} 
    \label{fig:overview}
\end{figure}

The core innovation of \method is our GAD framework, which synthesizes natural and adaptive behaviors from two complementary sources: a well-trained teacher policy and a reference motion dataset. As illustrated in Fig.~\ref{fig:overview}, GAD trains a student policy $\pi_\textrm{student}$ alongside two AMP-style discriminators, $\mathcal{D}_\textrm{teacher}$ and $\mathcal{D}_\textrm{dataset}$. Each discriminator evaluates the student's generated state transitions against one source of reference motions: either the teacher policy or the motion dataset.

Training proceeds in an interleaving manner, alternating between updating the student policy and the discriminators. In each iteration, we first update the student policy using the combined feedback from both discriminators and then train both discriminators to better distinguish between the student's outputs and their respective reference motions.

The teacher discriminator $\mathcal{D}_\textrm{teacher}$ optimizes:
$$
\begin{aligned}
\arg\min_{\mathcal{D}_\textrm{teacher}} & ~\mathbb{E}_{(s,s')\sim\pi_\textrm{teacher}}\left[\left(\mathcal{D}_\textrm{teacher}(s,s')-1\right)^2\right] \\
         & +\mathbb{E}_{(s,s')\sim\pi_\textrm{student}}\left[\left(\mathcal{D}_\textrm{teacher}(s,s')+1\right)^2\right] \\
         & +\lambda\mathbb{E}_{(s,s')\sim\pi_\textrm{teacher}}\left[\|\nabla_{(s,s')}\mathcal{D}_\textrm{teacher}(s,s')\|^2\right],
\end{aligned}
$$
while the reference discriminator $\mathcal{D}_\textrm{dataset}$ ensures natural motion qualities by optimizing:
$$
\begin{aligned}
\arg\min_{\mathcal{D}_\textrm{dataset}} & ~\mathbb{E}_{(s,s')\sim\mathcal{M}}\left[\left(\mathcal{D}_\textrm{dataset}(s,s')-1\right)^2\right] \\
         & +\mathbb{E}_{(s,s')\sim\pi_\textrm{student}}\left[\left(\mathcal{D}_\textrm{dataset}(s,s')+1\right)^2\right] \\
         & +\lambda\mathbb{E}_{(s,s')\sim\mathcal{M}}\left[\|\nabla_{(s,s')}\mathcal{D}_\textrm{dataset}(s,s')\|^2\right],
\end{aligned}
$$
where $\lambda$ controls the gradient penalty term that ensures stable training.

The student policy learns from a combined reward function:
$$r = r_\textrm{task} + w_\textrm{teacher} \cdot r_\textrm{teacher} + w_\textrm{dataset} \cdot r_\textrm{dataset},$$
where the discriminator rewards are computed as:
$$r_\textrm{teacher}=\max
\begin{bmatrix}
0, & 1-0.25(\mathcal{D}_\textrm{teacher}(s,s')-1)^2
\end{bmatrix}$$
$$r_\textrm{dataset}=\max
\begin{bmatrix}
0, & 1-0.25(\mathcal{D}_\textrm{dataset}(s,s')-1)^2
\end{bmatrix}$$

Both discriminators process state transitions using a consistent feature set comprising joint positions and velocities, root linear and angular velocities in the robot's local frame, base link orientation (roll and pitch), and root height. This common representation enables effective comparison across different motion sources while capturing the essential characteristics of locomotion behavior.

\begin{table*}[h]
\centering
\caption{Available observations in training}
\label{available_obs}
\resizebox{\textwidth}{!}{%
\begin{tabular}{lcccccccccccccc}
\toprule
\textbf{Sources} & \textbf{Phase} & \textbf{CmdVel} & \textbf{DoFPos} & \textbf{DoFVel} & \textbf{LastAction} & \textbf{Diff} & \textbf{BaseLinVel} & \textbf{BaseAngVel} & \textbf{RPY} & \textbf{Root Height} & \textbf{Push} & \textbf{Fraction} & \textbf{BodyMass} & \textbf{ContactStatus} \\
\midrule
Teacher & \checkmark & \checkmark & \checkmark & \checkmark & \checkmark & \checkmark & \checkmark & \checkmark & \checkmark &  & \checkmark & \checkmark & \checkmark & \checkmark \\
Dataset &  &  & \checkmark      & \checkmark      &        &        & \checkmark       & \checkmark       & \checkmark  &   \checkmark    &        &       &        &  \\
Student &      & \checkmark     & \checkmark      & \checkmark      & \checkmark       &        &        & \checkmark       & \checkmark       &       &        &        &  \\
\bottomrule
\multicolumn{15}{l}{\footnotesize \textit{Notes:}} \\
\multicolumn{15}{l}{\footnotesize $\bullet$ {Phase: Indicates the phase of motion, serving as a temporal marker.}} \\
\multicolumn{15}{l}{\footnotesize $\bullet$ {Diff: Difference between current joint angular position and reference joint angular position, calculated based on Phase.}} \\
\multicolumn{15}{l}{\footnotesize $\bullet$ {ContactStatus: Information regarding the stance mask and feet contact forces.}} \\
\end{tabular}%
}
\end{table*}
\textbf{Deployable Policy Distillation} A key aspect of our framework is enabling the student policy $\pi_\textrm{student}$ to generate actions when privileged observations are unavailable in real-world deployment. While the teacher policy benefits from privileged information during training to better understand task objectives and achieve efficient convergence, the student policy must learn to generate appropriate actions using only deployable sensor observations. This asymmetric approach allows us to leverage rich state information during training while ensuring the final policy remains deployable. The specific observations available to the student policy are detailed in Table~\ref{available_obs}.

\subsection{Training Process}
\textbf{Curriculum Learning}
Teacher policy $\pi_{teacher}$ training adopts a curriculum learning approach comprised of two distinct phases. The initial stability phase prioritizes maintaining balance and preventing falls, establishing fundamental stability behaviors. This is followed by the mobility phase, where the policy develops comprehensive omnidirectional locomotion capabilities. The specific reward components for each phase are detailed in Table \ref{teacher_reward}.

\textbf{Demonstration Data Preparation}
The locomotion motion data in this work is sourced from the LaFAN1 dataset and meticulously retargeted to conform to the kinematic specifications of Unitree H1 robots. While this dataset offers diverse motion styles and velocity ranges, utilizing all demonstrations simultaneously introduces ambiguity in the learning process. To facilitate distinct gait style demonstrations across different velocity commands, we strategically selected motion clips with minimal or non-overlapping velocity ranges, ensuring a relatively clear behavioral boundaries between different locomotion patterns.

\textbf{Asymmetric Actor-critic Architecture} 
Student policy training utilizes an asymmetric actor-critic architecture to effectively handle partial observability in real-world conditions. The student's observation processing begins with temporal partial observations $o^N_t = [o_{t-n}, o_{t-n+1} ... o_t]^T$. These observations are first processed through a partial states encoder $\mathcal{E}$ to generate context latent representations, which are then combined with the current partial state observations and the velocity command. The resulting combined representation passes through MLP layers to produce the final control actions.

\begin{table}[h]
\setlength{\abovecaptionskip}{0cm}
\setlength{\belowcaptionskip}{-0cm}
\caption{\small \textbf{Reward definitions used in Teacher policy training.}}
\label{teacher_reward}
\resizebox{\linewidth}{!}{
    \begin{tabular}{l|c|r}
    \toprule
    \textbf{Term} & \textbf{Definition} & \textbf{Weight} \\ \midrule
    \multicolumn{3}{c}{\textbf{First Stage}} \\ \midrule
    Termination &
    $\displaystyle r_{\text{termination}} = \mathbb{I}_{\mathrm{reset}}-\mathbb{I}_{\mathrm{timeout}}$
    & -1000 \\[2mm]
    Linear Velocity Tracking & $\exp\Bigl(-\frac{\| v_{xy}^{\text{target}}-v_{xy}\|_2}{0.1}\Bigr)$ & 10 \\
    Angular Velocity Tracking & $\exp\Bigl(-\frac{\|\omega_{z}^{\text{target}}-\omega_{z}\|_2}{0.1}\Bigr)$ & 10 \\[2mm]
    Linear Velocity $z$ & $\| v_z \|_2$ & -1.0 \\
    R-P Angular Velocity & $\|\omega_{xy}\|_2$ & -0.5 \\[2mm]
    Orientation & $\sum\limits_{i\in\{x,y\}} \bigl(\text{projected gravity}_i\bigr)^2$ & -1.0 \\[2mm]
    Base Height &
    $\displaystyle \exp\Bigl(-100\,\Bigl|h_{\text{base}} - h_{\text{target}}\Bigr|\Bigr) \newline
    where h_{\text{base}} = z_{\text{root}} - (h_{\text{feet}}-0.08)$
    & 0.5 \\[2mm]
    Action Rate & $\| a_t - a_{t-1} \|_2$ & -0.01 \\[2mm]
    Energy Square &
    $\displaystyle  \frac{\sum_{i=1}^{10} (\tau_i\,\dot{q}_i)^2}{1+\|\mathbf{c}_{xy}\|_2}$
    & -5e-6 \\[2mm]
    Stand Still &
    $\displaystyle \Bigl(\sum\bigl|q - q_{\text{default}})\cdot \mathbb{I}_{\text{stand}}$
    & -1 \\[2mm]
    Feet Clearance &
    $\displaystyle \sum_{i}\mathbb{I}\Bigl\{\bigl|h_{\text{feet},i} - h_{\text{target}}\bigr| < 0.01\Bigr\} \cdot \Bigl(1 - \text{gait phase}_i\Bigr)$
    & 2.5 \\[2mm]
    Feet Contact Number &
    $\displaystyle \operatorname{mean}\Bigl(\mathbb{I}_{\{\text{contact} = \text{stance mask}\}} - \mathbb{I}_{\{\text{contact} \neq \text{stance mask}\}}\Bigr)$
    & 1 \\[2mm]
    Default Joint Position & $\|q_{[1:2]} - q^{\text{default}}_{[1:2]}\|_2 + \|q_{[6:7]} - q^{\text{default}}_{[6:7]}\|_2$ & 0.5 \\[2mm]
    Action Smoothness & $\| a_{t-2} - 2a_{t-1} + a_t \|_2$ & -0.001 \\[2mm]
    Feet Slip & $1 - \sum_i \exp\Bigl(-\|v_{xy}^{\text{foot},i}\|_2\Bigr)$ & -0.05 \\[2mm]
    Reference Joint Position & 
    $\displaystyle \exp(-2\|q - q_{\text{ref}}\|_2) - 0.5\min(\|q - q_{\text{ref}}\|_2, 0.5)$
    & 10 \\[2mm]
    Pelvis-Ankle $y$ Distance & 
    $\displaystyle (\|y_{\text{pelvis\_pitch}} - y_{\text{ankle\_L}}\| + \|y_{\text{pelvis\_pitch}} - y_{\text{ankle\_R}}\|) \cdot \mathbb{I}_{\{|v_y| < 0.1\}}$ 
    & -5 \\[2mm]
    Upper Joint Constraints & 
    $\displaystyle \sum\|q_{[12:14]} - q^{\text{default}}_{[12:14]}\| + \sum\|q_{[16:18]} - q^{\text{default}}_{[16:18]}\| + \|q_{10} - q^{\text{default}}_{10}\|$
    & -5 \\[2mm]
    \midrule
    \multicolumn{3}{c}{\textbf{Second Stage}} \\ \midrule
    Joint Torque & $\|\tau\|_2$ & -2e-5 \\[2mm]
    Joint Acceleration & $\|\ddot{q}\|_2$ & -1e-6 \\[2mm]
    Feet Contact Forces &
    $\displaystyle \sum_{i}\max\Bigl(\|\text{contact force}_i\|_2 - F_{\max},0\Bigr)$
    & -0.01 \\
    Torque When Stand-Still & 
    $\displaystyle \sum\Bigl[ (\tau_t-\tau_{t-1})^2 + (\tau_t+\tau_{t-2}-2\tau_{t-1})^2\Bigr] \cdot \mathbb{I}_{\text{stand}}$ & -1e-3 \\[2mm]
    Body Pitch & 
    $\displaystyle \|\text{pitch} - 0.01\|$
    & -5 \\[2mm]
    Body Roll & 
    $\displaystyle \|\text{roll}\|$
    & -10 \\[2mm]
    Track Velocity Hard & 
    $\displaystyle \frac{e^{-10\|v_{xy}^{\text{target}} - v_{xy}\|} + e^{-10|\omega_z^{\text{target}} - \omega_z|}}{2} - 0.2(\|v_{xy}^{\text{error}}\| + |\omega_z^{\text{error}}|)$
    & 50 \\[2mm]
    Ankle Air Time & 
    $\displaystyle \sum_i(t_{\text{air},i} - 0.2) \cdot \mathbb{I}_{\text{first\_contact},i} \cdot -\mathbb{I}_{\text{stand\_still}}$
    & 100 \\[2mm]
    Ankle Limits & 
    $\displaystyle -\sum_{i\in\{4,9\}} \text{clip}(q_i - q_{\text{min},i}, 0) + \text{clip}(q_{\text{max},i} - q_i, 0)$
    & -200 \\ 
    \bottomrule
    \multicolumn{3}{l}{\footnotesize \textit{Notes:}} \\
    \multicolumn{3}{l}{\footnotesize $\bullet$ 
    \( \mathbb{I}_A = 1 \) if \( A=true \) and \( \mathbb{I}_A = 0 \) otherwise.} \\
    \multicolumn{3}{l}{\footnotesize $\bullet$ The maximum allowable feet contact force $F_{\text{max}}$ is set to 550N} \\[1mm]
    \end{tabular}
}

\vspace{-6pt}
\end{table}

\subsection{Implementation and Deployment Details}

Both policies are implemented using the Proximal Policy Optimization (PPO) algorithm \cite{schulman2017proximal}, with comprehensive domain randomization ensuring robust real-world transfer. 

\textbf{Domain Randomization}  
Following existing researches on humanoid whole-body control, our domain randomization encompasses three aspects: physical parameter variations, systematic observation noise injection, and randomized external force perturbations. The physical parameters include variations in mass distribution, joint properties, and surface interactions. Observation noise is carefully calibrated to match real-world sensor characteristics, while external forces simulate unexpected disturbances the robot might encounter during deployment.

\textbf{Safe Deployment} 
Safe deployment is achieved through torque limiting. This controller continuously monitors and adjusts torque outputs to remain within safe operational limits. The deployment architecture operates with the policy executing at 50Hz, while the low-level control loop maintains precise actuation at 1000Hz, ensuring responsive and stable behavior.

Real-world execution incorporates additional safety measures through continuous monitoring of joint positions, velocities, and torques. When approaching operational limits, the system smoothly modulates commands to maintain safe operation while preserving task performance. This approach enables robust deployment across varying conditions while protecting the hardware from potential damage.

\section{EXPERIMENTS}
We conduct comprehensive experiments in both simulation and real-world environments to evaluate \method's effectiveness in generating natural and adaptive locomotion behaviors. Our evaluation framework addresses four key aspects: (1) the effectiveness of GAD's distillation capabilities, (2) the accuracy of velocity tracking during locomotion tasks, (3) the quality of motion style reproduction, and (4) real-world deployment performance.

All experiments are conducted using the Unitree H1 humanoid robot in both simulated and physical environments. For reference motions, we utilize the LaFAN1 dataset, carefully retargeted to match the H1's kinematics. The motion data comprises global root position and orientation (quaternion), along with joint angular positions. Simulated experiments are performed in the NVIDIA Isaac Gym environment, which enables efficient parallel training and evaluation.

\subsection{Distillation Performance}
Our first set of experiments evaluates GAD's ability to effectively distill privileged information from the teacher policy while maintaining task performance. We compare GAD against several baseline distillation approaches, measuring both task achievement and motion naturalness. 

One of the main contributions of this work is the development of a Generative Adversarial Distillation method. In this context, we emphasize the ability of our single teacher discriminator (GAD-SD) to effectively distill knowledge from the teacher policy. To evaluate this capability, we compare our method against DAgger, one of the most widely used distillation methods in robot control.

First, we train an omnidirectional locomotion policy as the teacher. The command ranges used for both teacher training and the subsequent distillation experiment are listed in Table.~\ref{tab:loco_cmd_range}. We then leverage the well-trained teacher policy to guide the learning of the student policy.

\begin{table}[h]
\setlength{\abovecaptionskip}{0cm}
\setlength{\belowcaptionskip}{-0cm}
\centering
\caption{Ranges of Locomotion Task Command}
\label{tab:loco_cmd_range}
\resizebox{\columnwidth}{!}{%
    \begin{tabular}{lcccc} 
    \toprule
    \textbf{Parameter} & \makecell{Teacher (Unit)} & \makecell{Distillation \\student (Unit)} & \makecell{StyleLoco \\student (Unit)} \\
    \midrule
    % \multirow
      Forward (\(v_x\))   & \([-1.0, 3.5]\) \textit{m/s}& \([-1.0, 3.5]\) \textit{m/s} & \([-1.0, 4.5]\) \textit{m/s} & \\ 
      Lateral (\(v_y\))   & \([-0.8, 0.8]\) \textit{m/s} & \([-0.8, 0.8]\) \textit{m/s}&  \([-1.0, 1.0]\) \textit{m/s} & \\ 
      Angular (\(\omega_z\)) &  ~~\([-1.0, 1.0]\) \textit{rad/s} &  ~~\([-1.0, 1.0]\) \textit{rad/s}&  ~~\([-1.5, 1.5]\) \textit{rad/s} & \\ 
    \bottomrule
    \end{tabular}
}
\vspace{-0.2cm}
\end{table}

The evaluation metrics include linear velocity tracking reward, angular velocity tracking reward, and average survival time. As shown in Table~\ref{tab:distill}, while both methods successfully learn from the teacher policy, GAD-SD demonstrates superior performance, particularly in linear velocity tracking and survival time.

\begin{table}[h]
\centering
\caption{Quantitative comparison of distillation methods}
\label{tab:distill}
\begin{tabular}{lccccc}
\toprule
\textbf{Method} & \makecell{Linear Velocity\\Tracking\\ Reward(±0.1) $\uparrow$} & \makecell{Angular Velocity\\Tracking \\Reward(±0.1) $\uparrow$} & \makecell{Average Survival\\Time(±15 steps) $\uparrow$} \\
\midrule
Teacher          & 7.403 & 2.824 & 925.9 \\
\midrule
DAgger           & 3.744  & 2.516  & 506.6 \\
GAD-SD           & \textbf{5.679}  & \textbf{2.653}  & \textbf{860.3}\\
\bottomrule
\multicolumn{4}{l}{\footnotesize \textit{Notes:}} \\
\multicolumn{4}{l}{\footnotesize $\bullet$ Teacher: teacher policy trained with privileged information} \\
\multicolumn{4}{l}{\footnotesize $\bullet$ GAD-SD: GAD with only teacher distillation discriminator} \\
\end{tabular}
\end{table}

\subsection{Locomotion Capabilities}
The second set of experiments assesses the student policy's locomotion capabilities, particularly its ability to track commanded velocities while maintaining natural motion patterns. We compare \method against state-of-the-art approaches in terms of tracking accuracy, stability, and style preservation. Table~\ref{tab:comparison} shows comparative results across various performance metrics.

The locomotion task evaluates the ability of student policy to track local velocity commands comprising three components: forward/backward velocity $v_x$, lateral velocity $v_y$, and rotational velocity $w_z$. Command values are uniformly sampled within pre-defined ranges specified in Table.~\ref{tab:loco_cmd_range}. For style imitation, we select four representative motion clips as reference targets for the style discriminator, with their corresponding velocity profiles detailed in Table.~\ref{tab:vel_motion}.

\begin{table}[h]
\centering
\caption{Velocity Profiles for Motion Clips}
\label{tab:vel_motion}
\resizebox{\columnwidth}{!}{%
\begin{tabular}{lccc}
\toprule
\textbf{Vel Profiles} & \makecell{Forward \\ (\textit{m/s})} & \makecell{Lateral \\ (\textit{m/s})} & \makecell{Angular \\ (\textit{rad/s})} \\
\midrule
Slow Forward   & \([0.089,\, 1.205]\)  & \([-0.396,\, 0.188]\) & \([-1.734,\, 0.906]\)  \\
Medium Forward & \([0.884,\, 2.067]\)  & \([-0.563,\, 0.306]\) & \([-2.044,\, 1.963]\)  \\
Fast Forward   & \([2.438,\, 4.378]\)  & \([-1.166,\, 0.943]\) & \([-1.555,\, 3.476]\)  \\
Move Backward  & \([-1.088,\, -0.350]\) & \([-0.425,\, 0.365]\) & \([-1.580,\, 1.981]\)  \\
\bottomrule
\end{tabular}%
}
\end{table}

To comprehensively evaluate our double-discriminator framework, we compare our method against three baseline approaches:

\begin{itemize}
    \item SD-Motion: Single-discriminator approach using only motion clips as reference.
    \item SD-Full: Single-discriminator approach using a combination of teacher policy online roll-out data and motion clips.
    \item DAgger+Style: DAgger-based teacher policy distillation combined with a separate discriminator for style learning.
\end{itemize}

The evaluation metrics are similar to those used in the distillation task experiment, with the addition of energy consumption.

\begin{table*}[h]
\centering
\caption{Quantitative comparison of different methods across various metrics}
\label{tab:comparison}
\begin{tabular}{lccccc}
\toprule
\textbf{Method} & \makecell{Linear Velocity\\Tracking Reward(±0.1) $\uparrow$} & \makecell{Angular Velocity\\Tracking Reward(±0.1) $\uparrow$} & \makecell{Average Survival \\Time(±15 steps) $\uparrow$} & \makecell{Energy\\Consumption(±0.001) $\downarrow$} &\\
\midrule
SD-Motion        & 4.229 & 2.249 & 813.2 & \textbf{0.065}\\
SD-Full          & 4.665 & 2.413 & 824.1 & 0.093 \\
DAgger+Style     & 5.059 & 2.384 & 826.9 & 0.079 \\
GAD (Ours)       & \textbf{5.485} & \textbf{2.644} & \textbf{846.5} & 0.081 & \\
\bottomrule
\multicolumn{6}{l}{\footnotesize \textit{Notes:}} \\
\multicolumn{6}{l}{\footnotesize $\bullet$ SD-Motion: Single discriminator with only motion demonstrations} \\
\multicolumn{6}{l}{\footnotesize $\bullet$ SD-Full: Single discriminator with both teacher roll-outs and motion demonstrations} \\
\multicolumn{6}{l}{\footnotesize $\bullet$ DAgger+Style: DAgger distillation with additional style discriminator} \\
\end{tabular}
\end{table*}

As demonstrated in Table.~\ref{tab:comparison}, our proposed double-discriminator framework achieves superior performance in velocity tracking and survival time compared to all baseline methods.
Notably, the SD-Motion approach exhibits the best energy consumption performance, suggesting that human motions are inherently energy efficient and properly incorporating motion demonstrations during training contributes to reduced energy consumption.

\begin{figure}[h]
    \centering
    \includegraphics[width=0.9\linewidth]{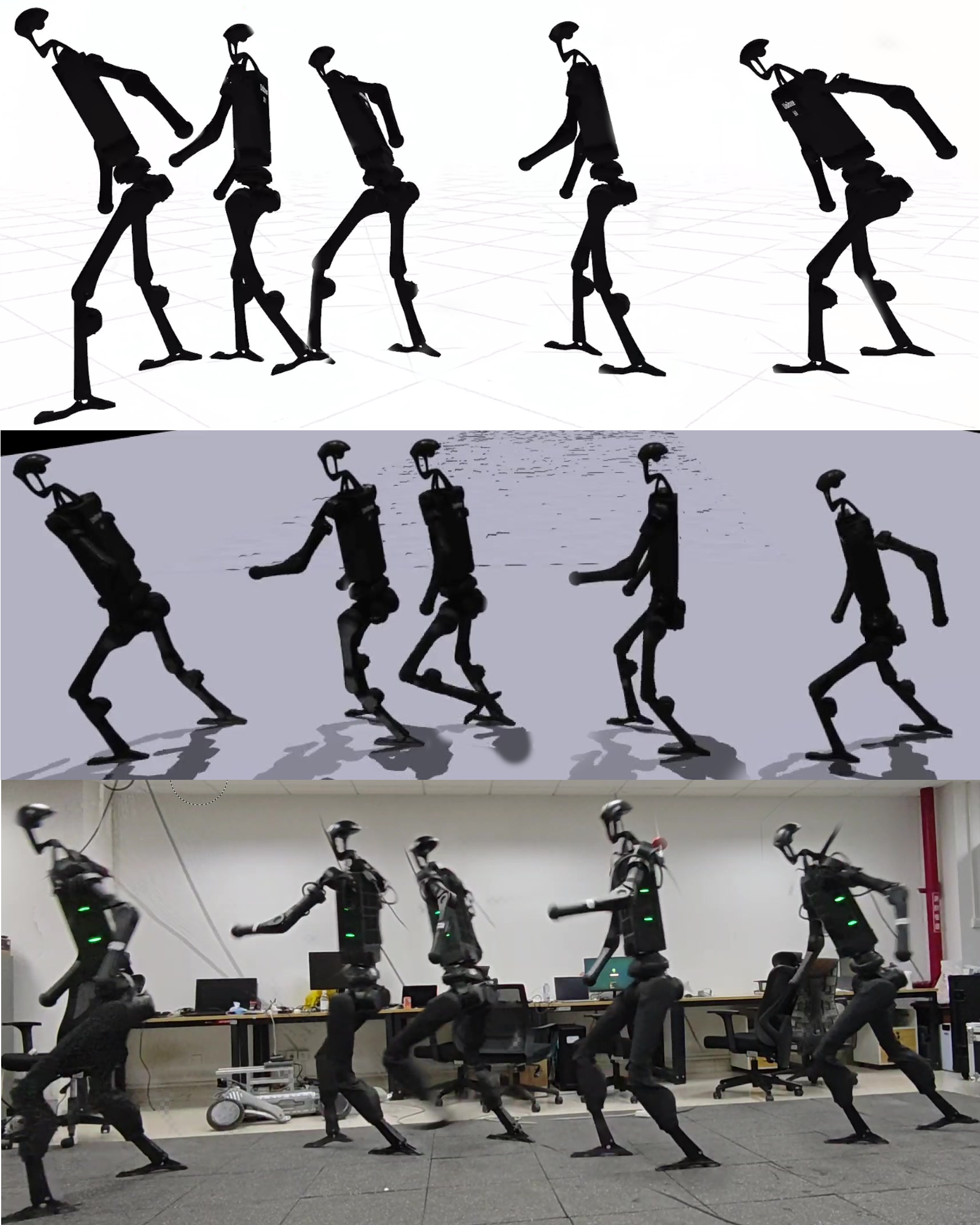}
    \caption{From top to bottom, a stylized locomotion demonstration from LaFAN1 (Top), motions generated by student policy in simulation (Middle), motions generated by student policy deployed on real H1 robot(Bottom).}
    \label{figs:stylized_locomotion}
\end{figure}

\subsection{Evaluations on Style Imitation}
To demonstrate our method's ability to combine robust locomotion skills with distinct motion styles, we evaluate a particularly challenging case: synthesizing a limping gait by combining a regular walking teacher policy with reference motions exhibiting a distinct limping pattern. Fig.~\ref{figs:stylized_locomotion} shows the comparison between the original limping motion from LaFAN1 (visualized in Rerun~\cite{RerunSDK}), the synthesized motion in Isaac Gym~\cite{makoviychuk2021isaacgymhighperformance}, and the deployed behavior on the physical Unitree H1 robot. The results demonstrate that our method successfully maintains the characteristic limping style while preserving the fundamental locomotion capabilities of the teacher policy.

This fusion of different motion sources creates an inherent trade-off between style fidelity and command tracking accuracy, as the stylized motions often deviate significantly from the teacher's optimal movement patterns. Our framework addresses this challenge through adjustable discriminator weights, allowing fine-tuned balance between style preservation and task performance.

\subsection{Real Robot Deployment}
\begin{figure}[h]
    \centering
    \includegraphics[width=0.8\linewidth]{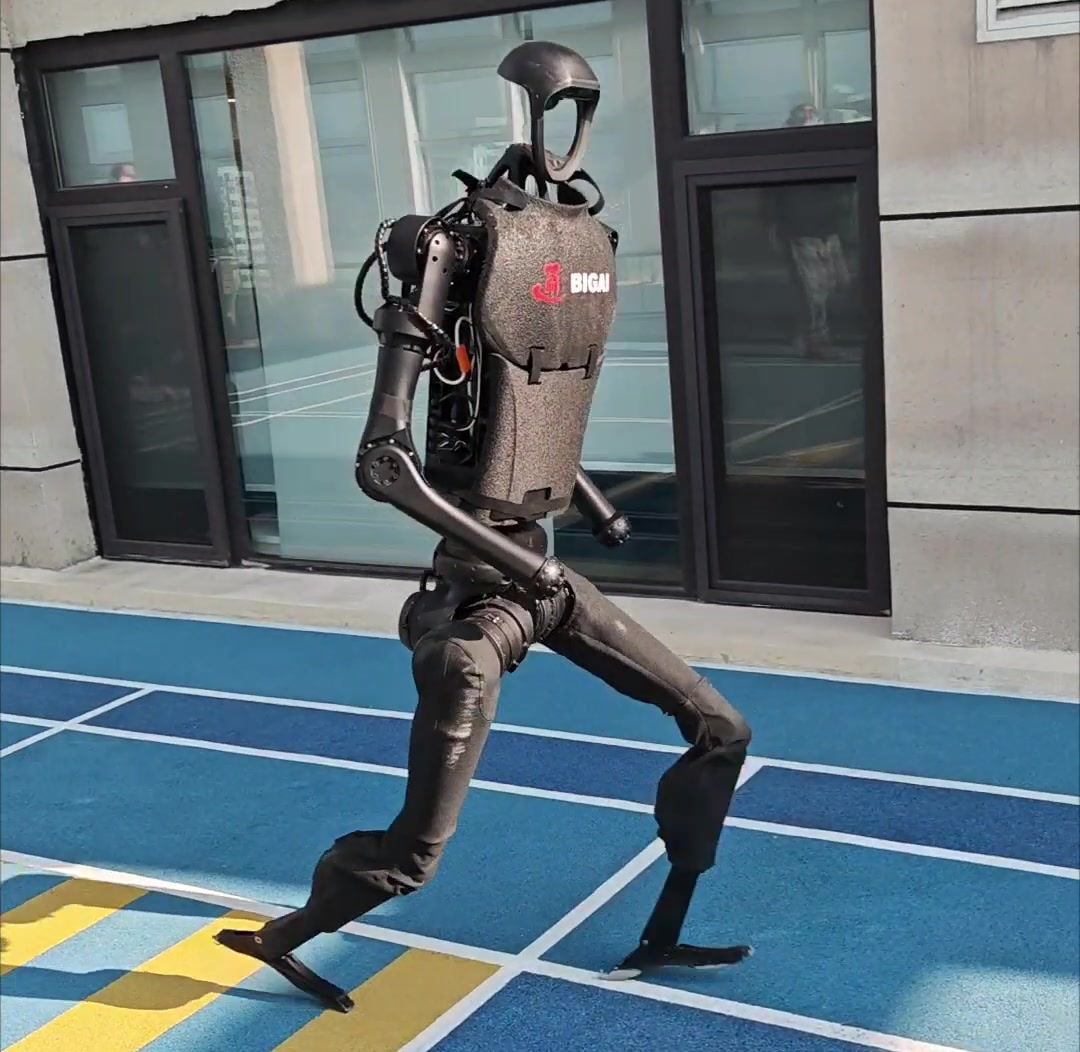}
    \caption{H1 operating outdoors at forward velocity ($v_{x}$) of 3 m/s}
    \label{figs:high_dynamics}
\end{figure}

The real-world deployment of our student policy on the Unitree H1 robot validates the practical effectiveness of our approach across various scenarios. As shown in Fig.~\ref{figs:gait_transition}, the robot demonstrates smooth transitions in both gait patterns and arm postures when responding to velocity command changes from low to medium speeds. The policy's robustness is further evidenced in Fig.~\ref{figs:high_dynamics}, where the robot maintains stable locomotion at high speeds up to 3 m/s. Most notably, Fig.~\ref{figs:stylized_locomotion} showcases our method's unique capability to synthesize stylized gaits that combine the stability of the teacher policy with distinctive motion patterns from the reference datasets, resulting in natural and controllable locomotion behaviors.

\section{CONCLUSION AND LIMITATIONS}
This paper presents \method, a novel framework for humanoid locomotion that bridges the gap between robust task execution and natural motion synthesis. Through our proposed Generative Adversarial Distillation approach, we demonstrate the effective combination of privileged information from expert policies with stylistic elements from human demonstrations. Our extensive experimental results, including successful deployment on the Unitree H1 robot, validate the framework's capability to generate stable and natural locomotion behaviors across diverse scenarios, from high-speed running at 3 m/s to stylized gaits such as limping.

The key innovation of our double-discriminator architecture enables simultaneous learning from heterogeneous sources while maintaining deployability through careful handling of privileged information. Quantitative evaluations show that \method outperforms existing approaches in both task performance and style preservation, demonstrating superior velocity tracking rewards and survival times while maintaining natural motion patterns.

Despite these achievements, several important limitations warrant future investigation. A primary challenge lies in style disambiguation when motion demonstrations share overlapping velocity ranges, potentially creating ambiguity in style selection and degrading imitation fidelity. Future research could explore automatic style clustering or context-aware selection mechanisms to address this limitation. Additionally, the current implementation relies on manual tuning of discriminator weights to balance task completion and style imitation objectives. Developing adaptive weighting schemes or automated tuning methods could enhance the framework's practical applicability. While our method shows impressive results in locomotion tasks, its generalization to broader manipulation tasks or more complex behaviors remains to be explored, opening avenues for future research.

Despite these limitations, \method represents a step toward natural and capable humanoid robotics, offering a promising foundation for future research in combining task-oriented control with human-like motion generation.

\bibliography{reference}
\bibliographystyle{ieeetran}
\end{document}